\documentclass[10pt,journal,letterpaper,compsoc,twocolumn]{IEEEtran}

\usepackage{cite}
\usepackage{graphicx}
\usepackage{psfrag}
\usepackage{subfigure}
\usepackage{url}
\usepackage{stfloats}
\usepackage{amssymb,amsmath}
\usepackage{amsfonts}
\usepackage{array}
\usepackage{CJK}
\usepackage{setspace}
\usepackage[ruled]{algorithm2e}
\usepackage{cases}
\usepackage{listings}

\begin{document}

\title{Residual Belief Propagation for Topic Modeling}

\author{Jia~Zeng,~\IEEEmembership{Member,~IEEE},
Xiao-Qin Cao
and
Zhi-Qiang~Liu
\IEEEcompsocitemizethanks{\IEEEcompsocthanksitem
J.~Zeng is with the School of Computer Science and Technology,
Soochow University, Suzhou 215006, China.
To whom correspondence should be addressed.
E-mail: j.zeng@ieee.org.
\IEEEcompsocthanksitem
X.-Q.~Cao and Z.-Q.~Liu are with the School of Creative Media,
City University of Hong Kong, Hong Kong, China.
}
}

\IEEEcompsoctitleabstractindextext{

\begin{abstract}
Fast convergence speed is a desired property for training latent Dirichlet allocation (LDA),
especially in online and parallel topic modeling for massive data sets.
This paper presents a novel residual belief propagation (RBP) algorithm
to accelerate the convergence speed for training LDA.
The proposed RBP uses an informed scheduling scheme for asynchronous message passing,
which passes fast-convergent messages with a higher priority to influence those slow-convergent messages at each learning iteration.
Extensive empirical studies confirm that RBP significantly reduces the training time until convergence while achieves a much lower predictive perplexity
than other state-of-the-art training algorithms for LDA,
including variational Bayes (VB),
collapsed Gibbs sampling (GS),
loopy belief propagation (BP),
and residual VB (RVB).
\end{abstract}

}

\maketitle

\section{Introduction} \label{s1}

Probabilistic topic modeling~\cite{Blei:12} is an important problem in machine learning and data mining.
As one of the simplest topic models,
latent Dirichlet allocation (LDA)~\cite{Blei:03} requires multiple iterations of training until convergence.
Recent studies find that
the convergence speed determines the efficiency of topic modeling for massive data sets.
For example,
online topic modeling algorithms~\cite{Hoffman:10} partition the entire data sets into mini-batches,
and optimize sequentially each mini-batch until convergence.
Another example lies in parallel topic modeling algorithms~\cite{Newman:09},
which optimize the distributed data sets until convergence and then synchronize the global topic distributions.
Therefore,
faster convergence speed leads to faster online and parallel topic modeling algorithms for massive data sets.

Training algorithms for LDA can be broadly categorized into variational Bayes (VB)~\cite{Blei:03},
collapsed Gibbs sampling (GS)~\cite{Griffiths:04} and loopy belief propagation (BP)~\cite{Zeng:11}.
According to a recent comparison~\cite{Zeng:11},
VB requires around $100$ iterations,
GS takes around $300$ iterations
and synchronous BP (sBP) needs around $170$ iterations to achieve convergence in terms of training perplexity.
Although VB uses the minimal number of iterations to achieve convergence,
its digamma function computation is so time-consuming as to slow down the convergence speed~\cite{Asuncion:09,Zeng:11}.
GS is a stochastic Markov chain Monte Carlo (MCMC) process,
which practically takes more iterations for convergence.
In contrast,
sBP is a deterministic scheme with smaller number of iterations until convergence than GS.
Moreover,
sBP does not involve complicated digamma functions,
and thus gains faster convergence speed over VB and GS.

In this paper,
we further adopt a residual belief propagation (RBP)~\cite{Elidan:06} algorithm to accelerate the convergence speed of topic modeling.
Compared with sBP,
RBP uses an informed scheduling strategy for asynchronous message passing,
in which it efficiently influences those slow-convergent messages by passing fast-convergent messages with a higher priority.
Through dynamically scheduling the order of message passing based on the residuals of two messages resulted from successive iterations,
RBP in theory converges significantly faster and more often than sBP~\cite{Elidan:06}.
The novelty of this paper is to introduce RBP into the probabilistic topic modeling community,
which significantly speeds up the convergence for training LDA.
Although jumping from sBP to RBP is a simple and straightforward idea,
extensive experimental results demonstrate that RBP in most cases converges fastest while reaches the lowest predictive perplexity
when compared with other state-of-the-art training algorithms,
including VB~\cite{Blei:03},
GS~\cite{Griffiths:04},
sBP~\cite{Zeng:11},
and residual VB (RVB)~\cite{Wahabzada:11,Wahabzada:11a}.
Because of its ease of use and fastest convergence speed,
RBP is a strong candidate for becoming the standard LDA training algorithm,
which may inspire faster online and parallel topic modeling algorithms in the near future.

\section{Related Work} \label{s2}

Recently,
LDA~\cite{Blei:03} has seen a rapid development for solving various topic modeling problems,
because of its elegant three-layer graphical representation as well as two efficient approximate inference methods like
VB~\cite{Blei:03} and GS~\cite{Griffiths:04}.
Both VB and GS have been widely used to learn LDA-based topic models until our recent work~\cite{Zeng:11}
reveals that there is yet another learning algorithm for LDA based on BP.
Extensive experiments show that the synchronous BP (sBP) is faster and more accurate than both VB and GS,
and has the potential to become a generic learning scheme for LDA-based topic models.
The basic idea of sBP is inspired by the factor graph~\cite{Kschischang:01} representation for LDA within the Markov random field (MRF) framework.
Similar BP ideas have been also proposed within the approximate mean-field framework~\cite{Asuncion:10}
as the zero-order approximation of collapsed VB (CVB0) algorithms~\cite{Asuncion:09}.

VB, GS and sBP can be explained within the unified message passing framework.
All these algorithms infer the marginal distribution of topic label for the word index called {\em message},
and estimate parameters by the iterative expectation-maximization (EM) algorithm according to the maximum-likelihood criterion~\cite{Bishop:book}.
They mainly differ in the E-step of EM algorithm for message update equations.
VB is a synchronous variational message passing algorithm~\cite{Winn:05},
which updates variational messages by complicated digamma functions,
introducing biases and slowing down the training speed~\cite{Asuncion:09,Zeng:11}.
GS updates messages by discrete topic labels randomly sampled from the message in the previous iteration.
Obviously,
the sampling process does not keep all uncertainties encoded in the previous message.
Also,
such a stochastic message updating often requires more iterations until convergence.
By contrast,
sBP directly uses the previous messages to update the current messages without sampling.
Such a deterministic process often takes the less number of iterations than GS to achieve convergence.

Similar to the proposed RBP,
residual VB (RVB) algorithms for LDA~\cite{Wahabzada:11,Wahabzada:11a}
have also been proposed from a matrix factorization perspective.
Because VB is in nature a synchronous message passing algorithm,
it does not have the direct asynchronous residual counterpart.
So,
RVB is derived from online VB (OVB) algorithms~\cite{Hoffman:10},
which divide the entire documents into mini-batches.
Through dynamically scheduling the order of mini-batches based on residuals,
RVB is often faster than OVB to achieve the same training perplexity.
Indeed,
there are several major differences between RVB and the proposed RBP.
First,
it is obvious that they are derived from different OVB and sBP algorithms, respectively.
While OVB can converge to the VB's objective function,
it practically involves complicated digamma functions for biases and the slowness~\cite{Asuncion:09,Zeng:11}.
Second,
RVB randomly generates a subset of mini-batches from a complicated residual distribution for training,
while the proposed RBP simply sorts residuals in a descending order for either documents or vocabulary words.
Notice that the random sampling process often misses those important mini-batches with largest residuals,
but the sorting technique ensures to locate those top documents or vocabulary words with largest residuals.
Because larger residuals correspond to more efficiency~\cite{Wahabzada:11,Wahabzada:11a},
our simple sorting technique in RBP is more efficient than the random sampling strategy in RVB.
This is one of the major reasons that RBP has a faster speed than RVB.
Finally,
RBP often achieves a much lower predictive perplexity than RVB,
partly because digamma functions in RVB introduce biases in parameter estimation.

\section{Residual Belief Propagation} \label{s3}

In this section,
we first introduce the conventional sBP algorithm for training LDA~\cite{Zeng:11}.
From the Markov random field (MRF) perspective,
the probabilistic topic modeling task can be interpreted as a labeling problem.
We assign a set of thematic topic labels,
$\mathbf{z} = \{z^k_{w,d}\}$,
to explain the nonzero elements in the document-word matrix,
$\mathbf{x} = \{x_{w,d}\}$,
where $1 \le w \le W$ and $1 \le d \le D$ are the word index in vocabulary and the document index in corpus.
The notation $1 \le k \le K$ is the topic index.
The nonzero element $x_{w,d} \ne 0$ denotes the number of word counts at the index $\{w,d\}$.
The topic label satisfies $\sum_k z^k_{w,d} = 1$,
where $z^k_{w,d} = \{0,1\}$.
To maximize the joint probability $p(\mathbf{x},\mathbf{z}|\alpha,\beta)$ of LDA,
the sBP algorithm computes the conditional marginal probability $p(z^k_{w,d}=1,x_{w,d}|\mathbf{z}^k_{-w,-d}, \mathbf{x}_{-w,-d})$,
called the {\em message},
$\mu(z^k_{w,d}=1) = \mu_{w,d}(k)$,
which can be normalized using a local computation, i.e.,
$\sum_{k=1}^K \mu_{w,d}(k) = 1, 0 \le \mu_{w,d}(k) \le 1$.
The message is proportional to the product of its neighboring messages,
\begin{align} \label{message}
\mu_{w,d}(k)\propto\frac{\boldsymbol{\mu}_{-w,d}(k) + \alpha}{\sum_k[\boldsymbol{\mu}_{-w,d}(k) + \alpha]} \times
&\frac{\boldsymbol{\mu}_{w,-d}(k) + \beta}{\sum_w[\boldsymbol{\mu}_{w,-d}(k) + \beta]},
\end{align}
where
\begin{gather}
\boldsymbol{\mu}_{-w,d}(k) = \sum_{-w} x_{-w,d}\mu_{-w,d}(k), \\
\boldsymbol{\mu}_{w,-d}(k) = \sum_{-d} x_{w,-d}\mu_{w,-d}(k),
\end{gather}
where $-w$ and $-d$ denote all word indices except $w$ and all document indices except $d$.
Based on messages,
the multinomial parameters $\theta$ and $\phi$ can be estimated as
\begin{gather}
\label{thetad}
\theta_d(k) = \frac{\boldsymbol{\mu}_{\cdot,d}(k) + \alpha}{\sum_k [\boldsymbol{\mu}_{\cdot,d}(k) + \alpha]}, \notag \\
\phi_w(k) = \frac{\boldsymbol{\mu}_{w,\cdot}(k) + \beta}{\sum_w [\boldsymbol{\mu}_{w,\cdot}(k) + \beta]}. \notag
\label{phiw}
\end{gather}
In sBP,
the synchronous schedule updates all messages simultaneously at iteration $t$ based on the messages at previous iteration $t-1$.
Although in practice this schedule often converges,
it often uses the more number of training iterations until convergence than VB~\cite{Zeng:11}.

The asynchronous schedule updates the message of each variable in a certain order,
which is in turn used to update other neighboring messages immediately at each iteration $t$.
The basic idea of RBP~\cite{Elidan:06} is to select the best updating order based on the messages' residuals $r_{w,d}$,
which are defined as the $p$-norm of difference between two message vectors at successive iterations,
\begin{align} \label{rbp1}
r_{w,d} = x_{w,d}\|\mu^t_{w,d} - \mu^{t-1}_{w,d}\|_p,
\end{align}
where $x_{w,d}$ is the number of word counts.
Here,
we choose the $L_1$ norm with $p=1$.
If we sequentially update message in a descending order of $r_{w,d}$ at each iteration,
the RBP algorithm theoretically converges faster or more often to a fixed point than sBP.
Because we extend sBP algorithms to classical RBP algorithms,
the theoretical proof of RBP's fast convergence rate remains the same as that in~\cite{Elidan:06}.

In practice,
the computational cost of sorting~\eqref{rbp1} is very high because we need to sort all non-zero residuals $r_{w,d}$ in the document-word matrix
at each learning iteration.
This scheduling cost is expensive in case of large-scale data sets.
Alternatively,
we may accumulate residuals based on either document or vocabulary indices,
\begin{gather}
\label{rbp2}
r_d = \sum_w r_{w,d}, \\
r_w = \sum_d r_{w,d}.
\label{rbp3}
\end{gather}
These residuals can be computed during message passing process at a negligible computational cost.
For large-scale data sets,
we advocate~\eqref{rbp3} because the vocabulary size is often a fixed number $W$ independent of the number of documents $D$.
So,
initially sorting $r_w$ requires at most a computational complexity of $\mathcal{O}(W\log W)$ using the standard quick sort algorithm.
If the successive residuals are in almost sorted order,
only a few swaps will restore the sorted order by the standard insertion sort algorithm,
thereby saving time.
In our experiments (not shown in this paper),
RBP based on~\eqref{rbp3} uses little computational cost to sort $r_w$
while retains almost the same convergence rate as that of sorting~\eqref{rbp1}.
We see that Eq.~\eqref{rbp2} is also useful for small-scale data sets,
because in this case $D < W$ as shown in Table~\ref{dataset}.

\begin{figure}[t]
\centering
\includegraphics[width=1\linewidth]{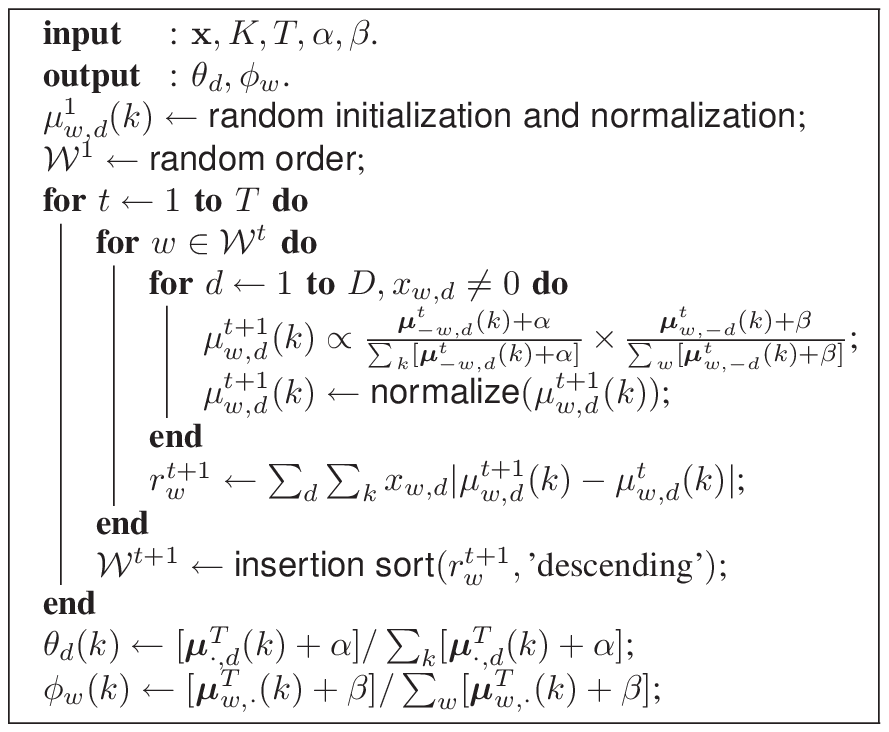}
\caption{The RBP algorithm for LDA.}
\label{rbp}
\end{figure}

Fig.~\ref{rbp} summarizes the proposed RBP algorithm based on~\eqref{rbp3},
which will be used in the following experiments.
First,
we initialize messages randomly and normalize them locally.
Second,
we start a random order of $w \in \mathcal{W}^1$ and accumulate residuals $r^{t+1}_w$ during message updating.
At the end of each learning iteration $t$,
we sort $r^{t+1}_w$ in the descending order to refine the updating order $w \in \mathcal{W}^{t+1}$.
Intuitively,
residuals reflect the convergence speed of message updating.
The larger residuals correspond to the faster-convergent messages.
In the successive learning iterations,
RBP always start passing fast-convergent messages with a higher priority in the order $\mathcal{W}^{t+1}$.
Because the asynchronous message passing influences the current message updating by the previous message updating,
passing fast-convergent messages will speed up the convergence of those slow-convergent messages.

\section{Experimental Results} \label{s4}

\begin{table}[t]
\caption{Statistics of six document data sets.}
\label{dataset}
\begin{center}
\begin{tabular}{|l|c|c|c|c|}
\hline
Data set   &$D$   &$W$   &$N_d$   &$W_d$     \\
\hline \hline
NG20        &$7505$     &$61188$    &$239$    &$129$     \\
BLOG        &$5177$     &$33574$    &$217$    &$149$     \\
CORA        &$2410$     &$2961$     &$57$     &$43$      \\
MEDLINE     &$2317$     &$8918$     &$104$    &$66$      \\
NIPS        &$1740$     &$13649$    &$1323$   &$536$     \\
WEBKB       &$7061$     &$2785$    &$50$    &$29$     \\
\hline
\end{tabular}
\end{center}
\end{table}

\begin{figure*}[t]
\centering
\includegraphics[width=0.8\linewidth]{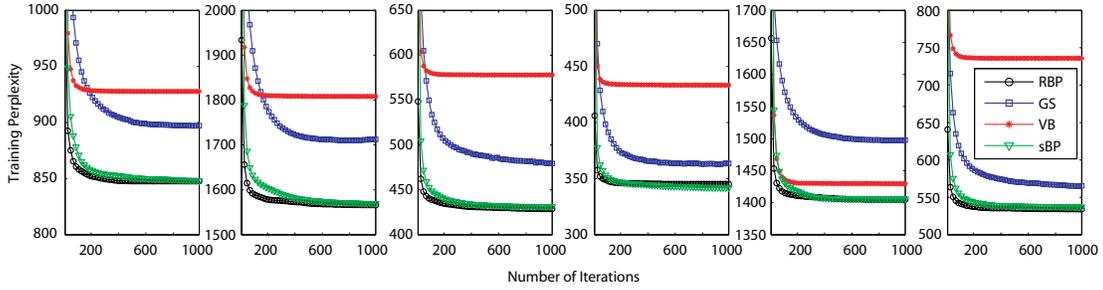}
\caption{Training perplexity as a function of number of iterations when $K=10$.}
\label{converge}
\end{figure*}

\begin{figure*}[t]
\centering
\includegraphics[width=0.8\linewidth]{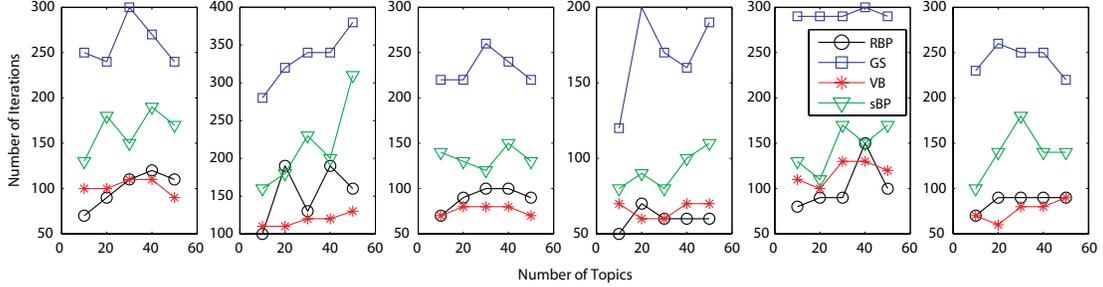}
\caption{Number of training iterations until convergence as a function of number of topics.}
\label{numcon}
\end{figure*}

\begin{figure*}[t]
\centering
\includegraphics[width=0.8\linewidth]{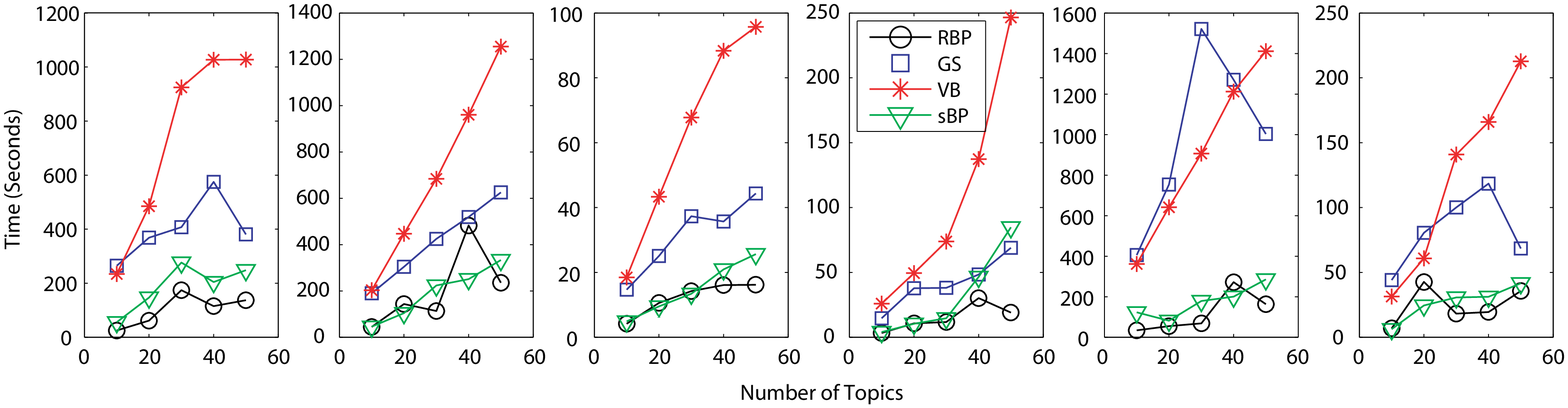}
\caption{Training time until convergence as a function of number of topics.}
\label{timecon}
\end{figure*}

\begin{figure*}[t]
\centering
\includegraphics[width=0.8\linewidth]{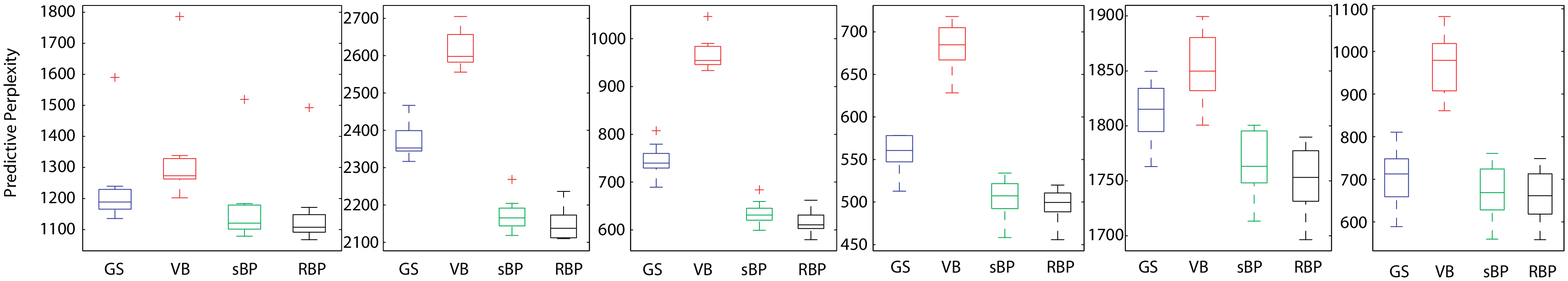}
\caption{Predictive perplexity for ten-fold cross-validation when $K=50$.}
\label{preperp}
\end{figure*}

\begin{figure*}[t]
\centering
\includegraphics[width=0.8\linewidth]{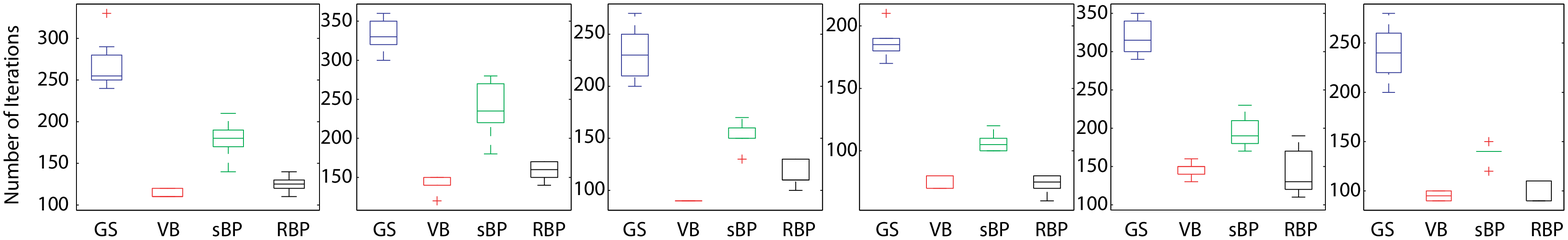}
\caption{Number of training iterations until convergence for ten-fold cross-validation when $K=50$.}
\label{prenumcon}
\end{figure*}

\begin{figure*}[t]
\centering
\includegraphics[width=0.8\linewidth]{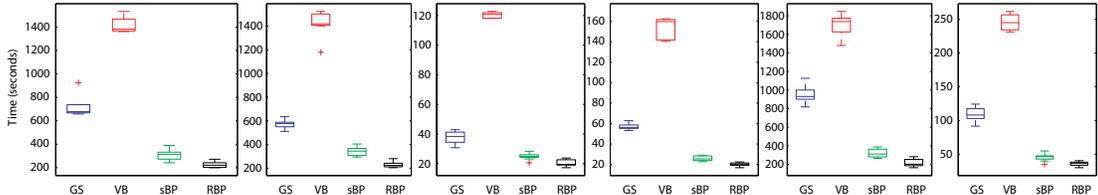}
\caption{Training time until convergence for ten-fold cross-validation when $K=50$.}
\label{pretimecon}
\end{figure*}

\begin{figure*}[t]
\centering
\includegraphics[width=0.7\linewidth]{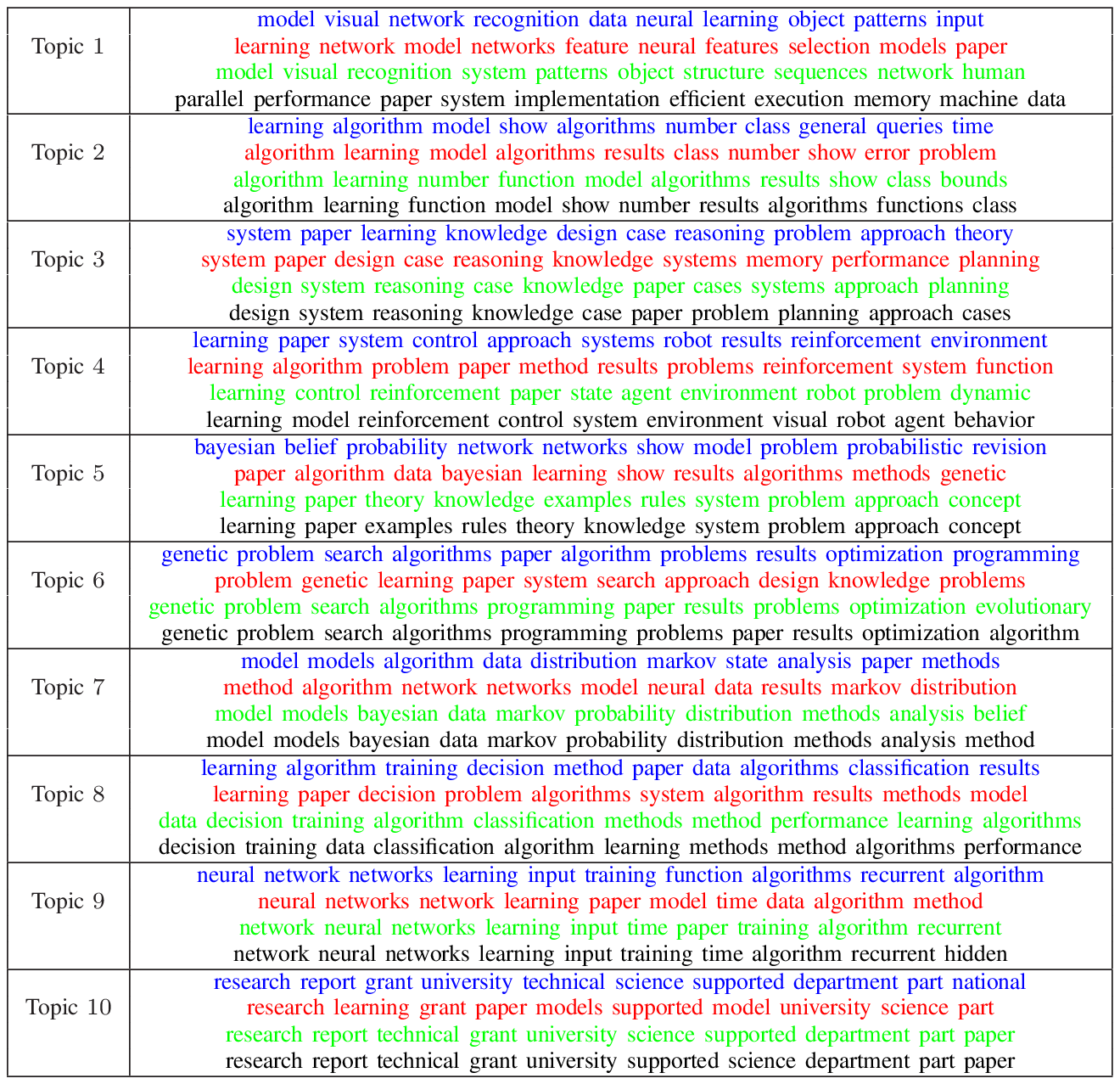}
\caption{Top ten words of $K=10$ topics for GS (blue), VB (red), sBP (green) and RBP (black) on CORA set.}
\label{topic}
\end{figure*}

We carry out experiments on six publicly available data sets:
1) 20 newsgroups (NG20),
2) BLOG,
3) CORA,
4) MEDLINE,
5) NIPS,
and 6) WEBKB.
Table~\ref{dataset} summarizes the statistics of six data sets,
where $D$ is the total number of documents in the corpus,
$W$ is the number of words in the vocabulary,
$N_d$ is the average number of word tokens per document,
and $W_d$ is the average number of word indices per document.
All subsequent figures show results on six data sets in the above order.
We compare RBP with three state-of-the-art approximate inference methods for LDA including VB~\cite{Blei:03},
GS~\cite{Griffiths:04},
and sBP~\cite{Zeng:11}
under the same fixed hyperparameters $\alpha = \beta = 0.01$.
We use MATLAB C/C++ MEX-implementations for all these algorithms~\cite{Zeng:12},
and carry out the experiments on a common PC with CPU $2.4$GHz and RAM $4$G.

Fig.~\ref{converge} shows the training perplexity~\cite{Asuncion:09} at every $10$ iterations in $1000$ iterations when $K=10$ for each data set.
All algorithms converge to a fixed point of training perplexity within $1000$ iterations.
Except the NIPS set,
VB always converges at the highest training perplexity.
In addition,
GS converges at a higher perplexity than both sBP and RBP.
While RBP converge at almost the same training perplexity as sBP,
it always reaches the same perplexity value faster than sBP.
Generally,
the training algorithm converges when the training perplexity difference at two consecutive iterations is below a threshold.
In this paper,
we set the convergence threshold to $1$
because the training perplexity decreases very little after this threshold is satisfied in Fig.~\ref{converge}.

Fig.~\ref{numcon} illustrates the number training iterations until convergence on each data set for different topics $K \in \{10,20,30,40,50\}$.
The number of iterations until convergence seems insensitive to the number of topics.
On the BLOG, CORA and WEBKB sets,
VB uses the minimum number iterations until convergence,
consistent with the previous results in~\cite{Zeng:11}.
For all data sets,
GS consumes the maximum number of iterations until convergence.
Unlike the deterministic message updating in VB, sBP and RBP,
GS uses the stochastic message updating scheme accounting for the largest number of iterations until convergence.
Although sBP costs significantly less number of iterations until convergence than GS,
it still uses much more number of iterations than VB.
By contrast,
through the informed dynamic scheduling for asynchronous message passing,
RBP on average converges more rapidly than sBP for all data sets.
In particular,
on the NG20, MEDLINE and NIPS sets,
RBP on average uses a comparable or even less number of iterations than VB until convergence.
Fig.~\ref{timecon} shows the training time in seconds until convergence on each data set for different topics $K \in \{10,20,30,40,50\}$.
Surprisingly,
while VB usually uses the minimum number iterations until convergence,
it often consumes the longest training time for these iterations.
The major reason may be attributed to the time-consuming digamma functions in VB,
which takes at least triple more time for each iteration than GS and sBP.
If VB removes the digamma functions,
it runs as fast as sBP.
Because RBP uses significantly less number of iterations until convergence than GS and sBP,
it consumes the least training time until convergence for all data sets in Fig.~\ref{timecon}.
We also examine the predictive perplexity of all algorithms until convergence based on a ten-fold cross-validation.
The predictive perplexity for the unseen test set is computed as that in~\cite{Asuncion:09}.
Fig.~\ref{preperp} shows the box plot of predictive perplexity for ten-fold cross-validation when $K=50$.
The plot produces a separate box for ten predictive perplexity values of each algorithm.
On each box,
the central mark is the median,
the edges of the box are the $25$th and $75$th percentiles,
the whiskers extend to the most extreme data points not considered outliers,
and outliers are plotted individually by the red plus sign.
Obviously,
VB yields the highest predictive perplexity,
corresponding to the worst generalization ability.
GS has a much lower predictive perplexity than VB,
but it has a much higher perplexity than both sBP and RBP.
The underlying reason is that GS samples a topic label from the messages
without retaining all possible uncertainties.
The residual-based scheduling scheme of RBP not only speeds up the convergence rate of sBP,
but also slightly lowers the predictive perplexity.
The reason is that RBP updates fast-convergent messages to efficiently influence those slow-convergent messages,
reaching fast to the local minimum of the predictive perplexity.
Figs.~\ref{prenumcon} and~\ref{pretimecon} illustrate the box plots for the number of iterations and the training time until convergence
for ten-fold cross-validation when $K=50$.
Consistent with Figs.~\ref{numcon} and~\ref{timecon},
VB consumes the minimum number of iterations,
but has the longest training time until convergence.
GS has the maximum of number of iterations,
but has the second longest training time until convergence.
Because RBP improves the convergence rate over sBP,
it consumes the least training time until convergence.

To measure the interpretability of inferred topics,
Fig.~\ref{topic} shows the top ten words of each topic when $K = 10$ on CORA set using $500$ training iterations.
We observe that both sBP and RBP can infer almost the same topics as other algorithms except the topic one,
where sBP identifies the ``pattern recognition" topic but RBP infers the ``parallel system" topic.
It seems that both sBP and RBP obtain slightly more interpretable topics than GS and VB especially in topic four,
where ``reinforcement learning" is closely related to ``control systems".
For other topics,
we find that they often share the similar top ten words but with different ranking orders.
More details on subjective evaluation for interpretability of topics can be found in~\cite{Chang:09b}.
However,
even if GS and VB yield comparably interpretable topics as RBP,
we still advocate RBP because it consumes less training time until convergence while reaches a much lower predictive perplexity value.

We also compare RBP with other residual-based techniques for training LDA such as RVB~\cite{Wahabzada:11,Wahabzada:11a}.
It is not easy to make a fair comparison because RBP is an offline learning but RVB is an online learning algorithm.
However,
using the same data sets WEBKB and NG20~\cite{Wahabzada:11},
we can approximately compare RBP with RVB using the training time when the predictive perplexity converges.
When $K=100$,
RVB converges at the predictive perplexity $600$ using $60$ seconds training time on WEBKB,
while it converges at the predictive perplexity $1050$ using $600$ seconds training time on NG20.
With the same experimental settings as RVB (hyperparameters $\alpha=\beta=0.01$),
RBP achieves the predictive perplexity $540$ using $35$ seconds for training on WEBKB,
while it achieves the predictive perplexity $1004$ using $420$ seconds for training on NG20.
The significant speedup is because RVB involves relatively slower digamma function computation,
and adopts a more complicated sampling method based on residual distributions for dynamic scheduling.

\section{Conclusions} \label{s5}

This paper presents a simple but effective RBP algorithm for training LDA.
Through the residual-based dynamic scheduling scheme,
RBP significantly improves the convergence rate of sBP but adding only an affordable scheduling cost for large-scale data sets.
On average,
it reduces around $50 \sim 100$ training iterations until convergence,
while achieves a relatively lower predictive perplexity than sBP.
For the ten-fold cross-validation on six publicly available document sets when $K=50$,
RBP on average reduces $63.7\%$ and $85.1\%$ training time until convergence than two widely-used GS and VB algorithms,
respectively.
Meanwhile,
it on average achieves $8.9\%$ and $22.1\%$ lower predictive perplexity than GS and VB,
respectively.
Compared with other residual techniques like RVB,
RBP reduces around $30\%\sim50\%$ training time to achieve the lower predictive perplexity.
While RBP is a simple extension of sBP~\cite{Zeng:11} by introducing the dynamic scheduling for message passing,
its theoretical basis~\cite{Elidan:06} and strong experimental results support its promising role
in the probabilistic topic modeling field.

\section*{Acknowledgements} \label{s6}

This work is substantially supported by NSFC (Grant No. 61003154),
the Shanghai Key Laboratory of Intelligent Information Processing,
China (Grant No. IIPL-2010-009),
and a grant from Baidu to JZ,
and a GRF grant from RGC UGC Hong Kong (GRF Project No.9041574)
and a grant from City University of Hong Kong (Project No. 7008026) to ZQL.

\bibliographystyle{IEEEtran}
\bibliography{IEEEabrv,RBP}

\begin{thebibliography}{10}
\providecommand{\url}[1]{#1}
\csname url@rmstyle\endcsname
\providecommand{\newblock}{\relax}
\providecommand{\bibinfo}[2]{#2}
\providecommand\BIBentrySTDinterwordspacing{\spaceskip=0pt\relax}
\providecommand\BIBentryALTinterwordstretchfactor{4}
\providecommand\BIBentryALTinterwordspacing{\spaceskip=\fontdimen2\font plus
\BIBentryALTinterwordstretchfactor\fontdimen3\font minus
  \fontdimen4\font\relax}
\providecommand\BIBforeignlanguage[2]{{%
\expandafter\ifx\csname l@#1\endcsname\relax
\typeout{** WARNING: IEEEtran.bst: No hyphenation pattern has been}%
\typeout{** loaded for the language `#1'. Using the pattern for}%
\typeout{** the default language instead.}%
\else
\language=\csname l@#1\endcsname
\fi
#2}}

\bibitem{Blei:12}
D.~M. Blei, ``Introduction to probabilistic topic models,''
  \emph{Communications of the ACM}.

\bibitem{Blei:03}
D.~M. Blei, A.~Y. Ng, and M.~I. Jordan, ``Latent {Dirichlet} allocation,''
  \emph{J. Mach. Learn. Res.}, vol.~3, pp. 993--1022, 2003.

\bibitem{Hoffman:10}
M.~Hoffman, D.~Blei, and F.~Bach, ``{Online learning for latent Dirichlet
  allocation},'' in \emph{NIPS}, 2010, pp. 856--864.

\bibitem{Newman:09}
D.~Newman, A.~Asuncion, P.~Smyth, and M.~Welling, ``Distributed algorithms for
  topic models,'' \emph{J. Mach. Learn. Res.}, vol.~10, pp. 1801--1828, 2009.

\bibitem{Griffiths:04}
T.~L. Griffiths and M.~Steyvers, ``Finding scientific topics,'' \emph{Proc.
  Natl. Acad. Sci.}, vol. 101, pp. 5228--5235, 2004.

\bibitem{Zeng:11}
J.~Zeng, W.~K. Cheung, and J.~Liu, ``Learning topic models by belief
  propagation,'' \emph{IEEE Trans. Pattern Anal. Mach. Intell.}, p.
  arXiv:1109.3437v4 [cs.LG], 2011.

\bibitem{Asuncion:09}
A.~Asuncion, M.~Welling, P.~Smyth, and Y.~W. Teh, ``{On smoothing and inference
  for topic models},'' in \emph{UAI}, 2009, pp. 27--34.

\bibitem{Elidan:06}
G.~Elidan, I.~McGraw, and D.~Koller, ``{Residual belief propagation: Informed
  scheduling for asynchronous message passing},'' in \emph{UAI}, 2006, pp.
  165--173.

\bibitem{Wahabzada:11}
M.~Wahabzada and K.~Kersting, ``Larger residuals, less work: Active document
  scheduling for latent {Dirichlet} allocation,'' in \emph{ECML/PKDD}, 2011,
  pp. 475--490.

\bibitem{Wahabzada:11a}
M.~Wahabzada, K.~Kersting, A.~Pilz, and C.~Bauckhage, ``More influence means
  less work: fast latent {Dirichlet} allocation by influence scheduling,'' in
  \emph{CIKM}, 2011, pp. 2273--2276.

\bibitem{Kschischang:01}
F.~R. Kschischang, B.~J. Frey, and H.-A. Loeliger, ``Factor graphs and the
  sum-product algorithm,'' \emph{IEEE Transactions on Inform. Theory}, vol.~47,
  no.~2, pp. 498--519, 2001.

\bibitem{Asuncion:10}
A.~Asuncion, ``{Approximate Mean Field for Dirichlet-Based Models},'' in
  \emph{ICML Workshop on Topic Models}, 2010.

\bibitem{Bishop:book}
C.~M. Bishop, \emph{Pattern recognition and machine learning}.\hskip 1em plus
  0.5em minus 0.4em\relax Springer, 2006.

\bibitem{Winn:05}
J.~Winn and C.~M. Bishop, ``Variational message passing,'' \emph{J. Mach.
  Learn. Res.}, vol.~6, pp. 661--694, 2005.

\bibitem{Zeng:12}
J.~Zeng, ``A topic modeling toolbox using belief propagation,'' \emph{J. Mach.
  Learn. Res.}, p. arXiv:1201.0838v1 [cs.LG], 2012.

\bibitem{Chang:09b}
J.~Chang, J.~Boyd-Graber, S.~Gerris, C.~Wang, and D.~Blei, ``Reading tea
  leaves: How humans interpret topic models,'' in \emph{NIPS}, 2009, pp.
  288--296.

\end{thebibliography}


\end{document}